%% file: main.tex
\documentclass[11pt,a4paper]{article}
\PassOptionsToPackage{linktocpage}{hyperref}
\usepackage[hyperindex,breaklinks]{hyperref}
\usepackage[hyperref]{naaclhlt2018}
\usepackage{times}
\usepackage{latexsym}
\usepackage{blindtext}
\usepackage{amsmath}
\usepackage{booktabs}
\usepackage{multirow}
\usepackage{hhline}
\usepackage{array}
\newcolumntype{T}{>{\tiny}l} 
\newcolumntype{t}{>{\tiny}c} 
\usepackage{tikz}
\usetikzlibrary{%
  trees,%
  shapes,%
  arrows,%
  positioning,%
  calc,%
  automata%
}
\usepackage{tikz-dependency}
\usepackage{pgfplots}
\pgfplotsset{compat=1.14}
\usepackage{booktabs}

\usepackage{url}

\aclfinalcopy 
\def\aclpaperid{15} 

\newcommand{\specialcell}[2][c]{%
  \begin{tabular}[#1]{@{}c@{}}#2\end{tabular}}

\newcommand\smalltt[1]{\texttt{\small #1}}
\newcommand\smallsc[1]{\textsc{\small #1}}

\definecolor{my_green}{rgb}{0, 0.6, 0.5}
\definecolor{my_orange}{rgb}{0.9, 0.6, 0}
\definecolor{my_purple}{rgb}{0.8, 0.6, 0.7}
\definecolor{my_blue}{rgb}{0, 0.45, 0.7}

\title{Olive Oil is Made \emph{of} Olives, Baby Oil is Made \emph{for} Babies:\\
Interpreting Noun Compounds using Paraphrases in a Neural Model}

\author{Vered Shwartz\thanks{~~Work done during an internship at Google.} \\
	Computer Science Department\\
	Bar-Ilan University\\
	Ramat-Gan, Israel\\
  {\tt vered1986@gmail.com} \\\And
  Chris Waterson \\
  Google Inc.\\
  1600 Amphitheatre Parkway\\
  Mountain View, California 94043\\
  {\tt waterson@google.com } \\}

\date{}

\begin{document}
\maketitle
\begin{abstract}
Automatic interpretation of the relation between the constituents of a noun compound, e.g. \textit{olive oil} (source) and \textit{baby oil} (purpose) is an important task for many NLP applications. Recent approaches are typically based on either noun-compound representations or paraphrases. While the former has initially shown promising results, recent work suggests that the success stems from memorizing single prototypical words for each relation. We explore a neural paraphrasing approach that demonstrates superior performance when such memorization is not possible.
\end{abstract}

\section{Introduction}
\label{sec:intro}
\input{introduction}

\section{Background}
\label{sec:background}
\input{background}

\section{Model}
\label{sec:model}
\input{model}

\section{Evaluation}
\label{sec:evaluation}
\input{evaluation}

\section{Analysis}
\label{sec:analysis}
\input{analysis}

\section{Conclusion}
\label{sec:conclusion}
\input{conclusion}

\section*{Acknowledgments}
We would like to thank Marius Pasca, 
Susanne Riehemann, Colin Evans, Octavian Ganea, and Xiang Li for the fruitful conversations, and 
Corina Dima for her help in running the compositional baselines.

\bibliography{bib}
\bibliographystyle{acl_natbib}

\newpage
\appendix
\input{technical_details}

\end{document}

%% file: introduction.tex
Automatic classification of a noun-compound (NC) to the implicit semantic relation that holds between its constituent words is beneficial for applications that require text understanding.
For instance, a personal assistant asked ``do I have a \textit{morning meeting} tomorrow?'' should search the calendar for meetings occurring in the morning, while for \textit{group meeting} it should look for meetings with specific participants. 
The NC classification task is a challenging one, as the meaning of an NC is often not easily derivable from the meaning of its constituent words \cite{sparck1983compound}. 

Previous work on the task falls into two main approaches. The first maps NCs to paraphrases that express the relation between the constituent words \cite[e.g.][]{nakov2006using,nulty2013general}, such as mapping \textit{coffee cup} and \textit{garbage dump} to the pattern $[w_1]\;\textsc{contains}\;[w_2]$. The second approach computes a representation for NCs from the distributional representation of their individual constituents. While this approach yielded promising results, recently, \newcite{W16-1604} showed that similar performance is achieved by representing the NC as a concatenation of its constituent embeddings, and attributed it to the \emph{lexical memorization} phenomenon \cite{levy-EtAl:2015:NAACL-HLT}.

In this paper we apply lessons learned from the parallel task of semantic relation classification. We adapt HypeNET \cite{shwartz-goldberg-dagan:2016:P16-1} to the NC classification task, using their path embeddings  to represent paraphrases and combining with distributional information. We experiment with various evaluation settings, including settings that make lexical memorization impossible. In these settings, the integrated method performs better than the baselines. Even so, the performance is mediocre for all methods, suggesting that the task is difficult and warrants further investigation.\footnote{The code is available at \href{https://github.com/tensorflow/models/tree/master/research/lexnet_nc}{https://github.com/tensorflow/} \href{https://github.com/tensorflow/models/tree/master/research/lexnet_nc}{models/tree/master/research/lexnet\_nc}.}

%% file: background.tex
Various tasks have been suggested to address noun-compound interpretation. NC paraphrasing extracts texts explicitly describing the implicit relation between the constituents, for example \textit{student protest} is a protest \textsc{led by}, \textsc{be sponsored by}, or \textsc{be organized by} students \cite[e.g.][]{nakov2006using,D11-1060,S13-2025,nulty2013general}. Compositionality prediction determines to what extent the meaning of the NC can be expressed in terms of the meaning of its constituents, e.g. \textit{spelling bee} is non-compositional, as it is not related to \textit{bee} \cite[e.g.][]{reddy-mccarthy-manandhar:2011:IJCNLP-2011}. In this paper we focus on the NC classification task, which is defined as follows: given a pre-defined set of relations, classify $nc = w_1 w_2$ to the relation that holds between $w_1$ and $w_2$. We review the various features used in the literature for classification.\footnote{Leaving out features derived from lexical resources \cite[e.g.][]{nastase2003exploring,tratz-hovy:2010:ACL}.} 

\subsection{Compositional Representations}
\label{sec:bg_compound_rep}

In this approach, classification is based on a vector representing the NC ($w_1~w_2$), which is obtained by applying a function to its constituents' distributional representations: $\vec{v}_{w_1}, \vec{v}_{w_2} \in \mathcal{R}^n$. Various functions have been proposed in the literature.

\newcite{mitchell2010composition} proposed 3 simple combinations of $\vec{v}_{w_1}$ and $\vec{v}_{w_2}$ (additive, multiplicative, dilation). Others suggested to represent compositions by applying linear functions, encoded as matrices, over word vectors. \newcite{baroni2010nouns} focused on adjective-noun compositions (AN) and represented adjectives as matrices, nouns as vectors, and ANs as their multiplication. Matrices were learned with the objective of minimizing the distance between the learned vector and the observed vector (computed from corpus occurrences) of each AN. The full-additive model \cite{zanzotto2010estimating,dinu-pham-baroni:2013:CVSC} is a similar approach that works on any two-word composition, multiplying each word by a square matrix: $nc = A \cdot \vec{v}_{w_1} + B \cdot \vec{v}_{w_2}$.

\newcite{D12-1110} suggested a non-linear composition model. A recursive neural network operates bottom-up on the output of a constituency parser to represent variable-length phrases. Each constituent is represented by a vector that captures its meaning and a matrix that captures how it modifies the meaning of constituents that it combines with. For a binary NC, $nc = g(W \cdot [\vec{v}_{w_1}; \vec{v}_{w_2}])$, where $W \in \mathcal{R}^{2n \times n}$ and $g$ is a non-linear function. 

These representations were used as features in NC classification, often achieving promising results \cite[e.g.][]{vandecruys-afantenos-muller:2013:SemEval-20132,dima2015automatic}. However, \newcite{W16-1604} recently showed that similar performance is achieved by representing the NC as a concatenation of its constituent embeddings, and argued that it stems from memorizing prototypical words for each relation. For example, classifying any NC with the head \textit{oil} to the \textsc{source} relation, regardless of the modifier. 

\subsection{Paraphrasing}
\label{sec:bg_paraphrasing}

In this approach, the paraphrases of an NC, i.e. the patterns connecting the joint occurrences of the constituents in a corpus, are treated as features. For example, both \textit{paper cup} and \textit{steel knife} may share the feature \textsc{made of}. \newcite{seaghdha2013interpreting} leveraged this ``relational similarity'' in a kernel-based classification approach. They combined the relational information with the complementary lexical features of each constituent separately. Two NCs labeled to the same relation may consist of similar constituents (\textit{paper}-\textit{steel}, \textit{cup}-\textit{knife}) and may also appear with similar paraphrases. Combining the two information sources has shown to be beneficial, but it was also noted that the relational information suffered from data sparsity: many NCs had very few paraphrases, and paraphrase similarity was based on ngram overlap. 

Recently, \newcite{surtani2015vsm} suggested to represent NCs in a vector space model (VSM) using paraphrases as features. These vectors were used to classify new NCs based on the nearest neighbor in the VSM. However, the model was only tested on a small dataset and performed similarly to previous methods.

%% file: model.tex
\begin{figure*}[!t]
\centering
\scalebox{.73}{\input{figures/architecture}}
\vspace*{-28pt}
\caption{An illustration of the classification models for the NC \textit{coffee cup}. The model consists of two parts: (1) the distributional representations of the NC (left, orange) and each word (middle, green). (2) the corpus occurrences of \textit{coffee} and \textit{cup}, in the form of dependency path embeddings (right, purple).} 
\vspace*{-12pt}
\label{fig:model}
\end{figure*}
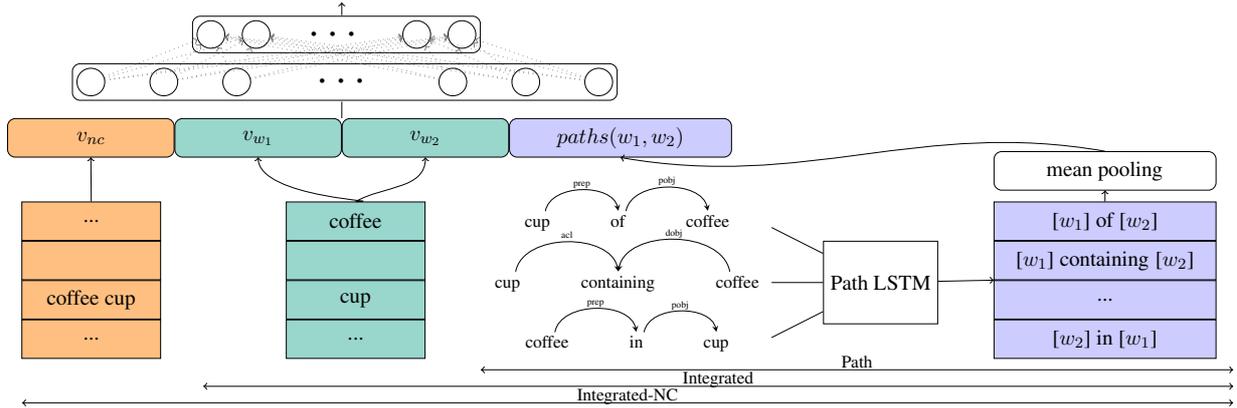

We similarly investigate the use of paraphrasing for NC relation classification. To generate a signal for the joint occurrences of $w_1$ and $w_2$, we follow the approach used by HypeNET \cite{shwartz-goldberg-dagan:2016:P16-1}. For an $w_1w_2$ in the dataset, we collect all the dependency paths that connect $w_1$ and $w_2$ in the corpus, and learn path embeddings as detailed in Section~\ref{sec:path_embeddings}. Section~\ref{sec:models} describes the classification models with which we experimented.

\subsection{Classification Models}
\label{sec:models}

Figure~\ref{fig:model} provides an overview of the models: path-based, integrated,
and integrated-NC, each which incrementally adds new features not present in the previous model.
In the following sections, $\vec{x}$ denotes the input vector representing the NC. The network classifies NC to the highest scoring relation: $r = \operatorname{argmax}_i \operatorname{softmax}(\vec{o})_i$, where $\vec{o}$ is the output layer. All networks contain a single hidden layer whose dimension is $\frac{|x|}{2}$. $k$ is the number of relations in the dataset. See Appendix~\ref{sec:technical_details} for additional technical details. 

\vspace{-5pt}
\paragraph{Path-based.} Classifies the NC based only on the paths connecting the joint occurrences of $w_1$ and $w_2$ in the corpus, denoted $P(w_1,w_2)$. We define the feature vector as the average of its path embeddings, where the path embedding $\vec{p}$ of a path $p$ is weighted by its frequency $f_{p, (w_1,w_2)}$:
\begin{equation*}
\vec{x} = \vec{v}_{P(w_1,w_2)} = \frac{\sum_{p \in P(w_1,w_2)} f_{p, (w_1,w_2)} \cdot \vec{p}}{\sum_{p \in P(w_1,w_2)} f_{p, (w_1,w_2)}}
\end{equation*}

\vspace{-5pt}
\paragraph{Integrated.} We concatenate $w_1$ and $w_2$'s word embeddings to the path vector, to add distributional information: $x = [\vec{v}_{w_1}, \vec{v}_{w_2}, \vec{v}_{P(w_1,w_2)}]$. Potentially, this allows the network to utilize the contextual properties of each individual constituent, e.g. assigning high probability to \textsc{substance-material-ingredient} for edible $w_1$s (e.g. \textit{vanilla pudding, apple cake}).

\vspace{-5pt}
\paragraph{Integrated-NC.} We add the NC's \emph{observed} vector $\vec{v}_{nc}$ as additional distributional input, providing the contexts in which $w_1w_2$ occur as an NC:\\ $\vec{v}_{nc} = [\vec{v}_{w_1}, \vec{v}_{w_2}, \vec{v}_{nc}, \vec{v}_{P(w_1,w_2)}]$. Like \newcite{W16-1604}, we learn NC vectors using the GloVe algorithm \cite{pennington-socher-manning:2014:EMNLP2014}, by replacing each NC occurrence in the corpus with a single token.

This information can potentially help clustering NCs that appear in similar contexts despite having low pairwise similarity scores between their constituents. 
For example, \textit{gun violence} and \textit{abortion rights} belong to the \textsc{topic} relation and may appear in similar news-related contexts, while \textit{(gun, abortion)} and \textit{(violence, rights)} are dissimilar.

\subsection{Path Embeddings}
\label{sec:path_embeddings}

Following HypeNET, for a path $p$ composed of edges $e_1,...,e_k$, we represent each edge by the concatenation of its lemma, part-of-speech tag, dependency label and direction vectors: $\vec{v_e} = [\vec{v_l}, \vec{v_{pos}}, \vec{v_{dep}}, \vec{v_{dir}}]$. The edge vectors $\vec{v_{e_1}}, ..., \vec{v_{e_k}}$ are encoded using an LSTM \cite{hochreiter1997long}, and the last output vector $\vec{p}$ is used as the path embedding. 

We use the NC labels as distant supervision. While HypeNET predicts a word pair's label from the frequency-weighted average of the path vectors, we differ from it slightly and compute the label from the frequency-weighted average of the predictions obtained from each path separately:
\begin{align*}
	& \vec{o} = \frac{\sum_{p \in P(w_1,w_2)} f_{p, (w_1,w_2)} \cdot \operatorname{softmax}(\vec{p})}{\sum_{p \in P(w_1,w_2)} f_{p, (w_1,w_2)}} \\ 
    & r = \operatorname{argmax}_i \vec{o}_i 
\end{align*}
\noindent We conjecture that label distribution averaging allows for more efficient training of path embeddings when a single NC contains multiple paths. 

%% file: figures/architecture.tex
\begin{tikzpicture}[
  normal/.style={rectangle,draw,fill=white,minimum width=4cm,minimum height=0.7cm,rounded corners=.8ex},
  path_embedding/.style={rectangle,draw,fill=blue!20,minimum width=4cm,    minimum height=0.7cm},
  nc_embedding/.style={rectangle,draw,fill=orange!50,minimum width=2.5cm,    minimum height=0.7cm},
  word_embedding/.style={rectangle,draw,fill=my_green!40,minimum width=2.5cm,    minimum height=0.7cm},
  every neuron/.style={circle,draw,minimum size=0.5cm},
  neuron missing/.style={draw=none,scale=2,execute at begin node=\color{black}$\cdots$}
  ]
  
    \node[nc_embedding] (nc1) {...};
    \node[nc_embedding] (nc2) [below=0 of nc1] {};
    \node[nc_embedding] (nc3) [below=0 of nc2] {coffee cup};
    \node[nc_embedding] (nc4) [below=0 of nc3] {...};
    
    \node[word_embedding] (w1) [right=2.25 cm of nc1] {coffee};
    \node[word_embedding] (w2) [below=0 of w1] {};
    \node[word_embedding] (w3) [below=0 of w2] {cup};
    \node[word_embedding] (w4) [below=0 of w3] {...};
    
    \node[path_embedding] (p1) [right=15 cm of nc1] {[$w_1$] of [$w_2$]};
    \node[path_embedding] (p2) [below=0 of p1] {[$w_1$] containing [$w_2$]};
    \node[path_embedding] (p3) [below=0 of p2] {...};
    \node[path_embedding] (p4) [below=0 of p3] {[$w_2$] in [$w_1$]};
    
\node[word_embedding,rounded corners=.8ex,minimum width=3cm] (v_w2) [above=0.8 of w1.north east] {$v_{w_2}$};
\node[word_embedding,rounded corners=.8ex,minimum width=3cm] (v_w1) [left=0 cm of v_w2] {$v_{w_1}$};
\node[nc_embedding,rounded corners=.8ex,minimum width=3cm] (v_w1_w2) [left=0 cm of v_w1] {$v_{nc}$};
    \node[path_embedding,rounded corners=.8ex] (p_w1_w2) [right=0 of v_w2] {$paths(w_1,w_2)$};
    \node[normal] (mean_pooling) [above=0.2 cm of p1] {mean pooling};
     \path[draw,->] (p1.north) -- (mean_pooling.south);
     \path[draw,->] (mean_pooling.north) to [out=170,in=-10] (p_w1_w2.south);
     \path[draw,->] (nc1.north) -- (v_w1_w2.south);
     \path[draw,->] (w1.north) to [out=0,in=-90] (v_w1.south);
     \path[draw,->] (w1.north) to [out=30,in=-90] (v_w2.south);
    
   \node [every neuron/.try, neuron 1/.try] (input-0) [above=0.4 cm of v_w1_w2] {};
   
   \def\lasty{0}
   \foreach \m/\l [count=\y,remember=\y as \lasty] in {2,3,missing,4,5,6}
  	\node [every neuron/.try, neuron \m/.try] (input-\y) [right=0.8 cm of input-\lasty]  {};

	\node [every neuron/.try, neuron 1/.try] (output-0) [above right=0.5 cm and 1.8 cm of input-0] {};
    \def\lasty{0}
   \foreach \m/\l [count=\y,remember=\y as \lasty] in {2,missing,3,4}
  	\node [every neuron/.try, neuron \m/.try] (output-\y) [right=0.3 cm of output-\lasty]  {};
    
    \draw[rounded corners=.8ex] ($(input-0.north west)+(-0.15,0.15)$) rectangle ($(input-6.south east)+(0.15,-0.15)$);
    
    \draw[rounded corners=.8ex] ($(output-0.north west)+(-0.15,0.15)$) rectangle ($(output-4.south east)+(0.15,-0.15)$);
    
    \node[draw=none] (output) [above=2.1 cm of v_w2.north west] {};
    \node[draw=none] (middle) [above=1.6 cm of v_w2.north west] {};
    \node[draw=none] (input) [above=0.3 cm of v_w2.north west] {};
    \draw (v_w2.north west) -- (input);
    \draw[->] (middle) -- (output);
    
  \foreach \i in {0,1,2,4,5,6}
    \foreach \j in {0,1,3,4}
      \draw [gray,dotted,->] (input-\i) -- (output-\j);

\node[normal,rounded corners=0ex,minimum height=1.5cm,minimum width=1 cm] (path_lstm) [below left=0 cm and 1 cm of p1] {Path LSTM};
\draw[->] (path_lstm) -- (p2.south west);

\begin{scope}[shift={($(path_lstm.east)+(-5.6cm,1.1cm)$)}]
\footnotesize
  \begin{deptext}[column sep=3em]
    cup \& of \& coffee \\
  \end{deptext}
  \depedge[theme=simple]{1}{2}{prep}
  \depedge[theme=simple]{2}{3}{pobj}
\end{scope}  

\begin{scope}[shift={($(path_lstm.east)+(-5.6cm,0cm)$)}]
\footnotesize
  \begin{deptext}[column sep=3em]
    cup \& containing \& coffee  \\
  \end{deptext}
  \depedge[theme=simple]{1}{2}{acl}
  \depedge[theme=simple]{3}{2}{dobj}
\end{scope}

\begin{scope}[shift={($(path_lstm.east)+(-5.6cm,-1.1cm)$)}]
\footnotesize
  \begin{deptext}[column sep=3em]
    coffee \& in \& cup  \\
  \end{deptext}
  \depedge[theme=simple]{1}{2}{prep}
  \depedge[theme=simple]{2}{3}{pobj}
\end{scope}

\draw ($(path_lstm.east)+(-3cm,0cm)$) -- (path_lstm);
\draw ($(path_lstm.east)+(-3cm,1cm)$) -- (path_lstm);
\draw ($(path_lstm.east)+(-3cm,-1cm)$) -- (path_lstm);

\draw[<->] ($(p4.south east)+(0.3,-0.2)$) -- ($(w4.south east)+(1,-0.2)$) node [midway,yshift=0.15cm] {\small Path};
\draw[<->] ($(p4.south east)+(0.3,-0.5)$) -- ($(w4.south east)+(-4,-0.5)$) node [midway,yshift=0.13cm] {\small Integrated};
\draw[<->] ($(p4.south east)+(0.3,-0.8)$) -- ($(nc4.south east)+(-2.5,-0.8)$) node [midway,yshift=0.12cm] {\small Integrated-NC};

\end{tikzpicture}

%% file: evaluation.tex
\begin{table*}[!t]
\small
\center
    \input{figures/results}
\vspace*{-7pt}
\caption{All methods' performance ($F_1$) on the various splits: \textbf{best freq}: best performing frequency baseline (head / modifier),\footnotemark~
\textbf{best comp}: best model from \newcite{W16-1604}.}
\label{tab:results}
\vspace*{-8pt}
\end{table*}

\begin{table}[!t]
\scriptsize
\center
\input{figures/splits}
\vspace{-7pt}
\caption{Number of instances in each dataset split.}
\label{tab:splits}
\vspace{-10pt}
\end{table}

\subsection{Dataset}
\label{sec:datasets}

We follow \newcite{W16-1604} and evaluate on the \newcite{tratz2011semantically} dataset, with 19,158 instances and two levels of labels: fine-grained (\smalltt{Tratz-fine}, 37 relations) and coarse-grained (\smalltt{Tratz-coarse}, 12 relations). We report results on both versions. See \newcite{tratz2011semantically} for the list of relations.

\paragraph{Dataset Splits}~\newcite{W16-1604} showed that a classifier based only on $v_{w_1}$ and $v_{w_2}$ performs on par with compound representations, and that the success comes from lexical memorization \cite{levy-EtAl:2015:NAACL-HLT}: memorizing the majority label of single words in particular slots of the compound (e.g. \textsc{topic} for \textit{travel guide}, \textit{fishing guide}, etc.). 
This memorization paints a skewed picture of the state-of-the-art performance on this difficult task.

To better test this hypothesis, we evaluate on 4 different splits of the datasets to train, test, and validation sets: (1) \textbf{random}, in a 75:20:5 ratio, (2) \textbf{lexical-full}, in which the train, test, and validation sets each consists of a distinct vocabulary. The split was suggested by \newcite{levy-EtAl:2015:NAACL-HLT}, and it randomly assigns words to distinct sets, such that for example, including \textit{travel guide} in the train set promises that \textit{fishing guide} would not be included in the test set, and the models do not benefit from memorizing that the head \textit{guide} is always annotated as \textsc{topic}. Given that the split discards many NCs, we experimented with two additional splits: (3) \textbf{lexical-mod} split, in which the $w_1$ words are unique in each set, and (4) \textbf{lexical-head} split, in which the $w_2$ words are unique in each set. Table~\ref{tab:splits} displays the sizes of each split.

\begin{table*}[!t]
\small
\center
\input{figures/indicative_paths}
\vspace{-7pt}
\caption{Indicative paths for selected relations, along with NC examples.}
\label{tab:indicative_paths}
\vspace{-8pt}
\end{table*}

\subsection{Baselines}
\label{sec:baselines}

\paragraph{Frequency Baselines.} \emph{mod freq} classifies $w_1w_2$ to the most common relation in the train set for NCs with the same modifier ($w_1w'_2$), while \emph{head freq} considers NCs with the same head ($w'_1w_2$).\footnotetext{In practice, in lexical-full this is a random baseline, in lexical-head it is the modifier frequency baseline, and in lexical-mod it is the head frequency baseline.}\footnote{Unseen heads/modifiers are assigned a random relation.}

\paragraph{Distributional Baselines.} Ablation of the path-based component from our models: \textbf{Dist} uses only $w_1$ and $w_2$'s word embeddings: $\vec{x} = [\vec{v}_{w_1}, \vec{v}_{w_2}]$, while \textbf{Dist-NC} includes also the NC embedding: $\vec{x} = [\vec{v}_{w_1}, \vec{v}_{w_2}, \vec{v}_{nc}]$. The network architecture is defined similarly to our models (Section~\ref{sec:models}).

\paragraph{Compositional Baselines.} We re-train Dima's \shortcite{W16-1604} models, various combinations of NC representations \cite{zanzotto2010estimating,D12-1110} and single word embeddings in a fully connected network.\footnote{We only include the compositional models, and omit the ``basic'' setting which is similar to our Dist model. For the full details of the compositional models, see \newcite{W16-1604}.}

\subsection{Results}
\label{sec:results}

Table~\ref{tab:results} shows the performance of various methods on the datasets. Dima's \shortcite{W16-1604} compositional models perform best among the baselines, and on the random split, better than all the methods. On the lexical splits, however, the baselines exhibit a dramatic drop in performance, and are outperformed by our methods. The gap is larger in the lexical-full split. Finally, there is usually no gain from the added NC vector in Dist-NC and Integrated-NC.

%% file: figures/results.tex
\begin{tabular}{c c c c c c c c c c }
\toprule
\textbf{Dataset} & \textbf{Split} & \textbf{Best Freq} &  \textbf{Dist} & \textbf{Dist-NC} & \textbf{Best Comp} & \textbf{Path} & \textbf{Int} & \textbf{Int-NC} \\ 
\hline
\multirow{4}{*}{\texttt{Tratz-fine}}
& Rand$_{}$ & 0.319 & 0.692 & 0.673 & \textbf{0.725} & 0.538 & 0.714 & 0.692 \\
& Lex$_{head}$ & 0.222 & 0.458 & 0.449 &  0.450 & 0.448 & \textbf{0.510} & 0.478 \\
& Lex$_{mod}$ & 0.292 & 0.574 & 0.559 & 0.607 & 0.472 & \textbf{0.613} & 0.600 \\
& Lex$_{full}$ & 0.066 & 0.363 &  0.360 &  0.334 & 0.423 & 0.421 & \textbf{0.429} \\
\midrule
\multirow{4}{*}{\texttt{Tratz-coarse}}
& Rand$_{}$ & 0.256 &  0.734 &  0.718 &  \textbf{0.775} & 0.586 &  0.736 & 0.712 \\
& Lex$_{head}$ & 0.225 & 0.501 & 0.497 & 0.538 & 0.518 &  \textbf{0.558} & 0.548 \\
& Lex$_{mod}$ & 0.282 &  0.630 &  0.600 & 0.645 &  0.548 & \textbf{0.646} &  0.632 \\
& Lex$_{full}$ & 0.136 &  0.406 & 0.409 & 0.372 & 0.472 & 0.475 & \textbf{0.478} \\
\bottomrule
\end{tabular}

%% file: figures/splits.tex
\begin{tabular}{c c c c c}
\toprule
\textbf{Dataset} & \textbf{Split} &   \textbf{Train} & \textbf{Validation}   &  \textbf{Test} \\ 
\midrule
\multirow{4}{*}{\specialcell{\textsc{Tratz-fine}}} & Lex$_{full}$ &  4,730 & 1,614 & 869 \\
& Lex$_{head}$ &  9,185 & 5,819 & 4,154 \\
& Lex$_{mod}$ &  9,783 & 5,400 & 3,975 \\
& Rand &  14,369 & 958 & 3,831 \\
\midrule
\multirow{4}{*}{\specialcell{\textsc{Tratz-coarse}}} & Lex$_{full}$ & 4,746 & 1,619 & 779 \\
& Lex$_{head}$ & 9,214 & 5,613 & 3,964 \\
& Lex$_{mod}$ & 9,732 & 5,402 & 3,657 \\
& Rand &  14,093 & 940 & 3,758 \\
\bottomrule
\end{tabular}

%% file: figures/indicative_paths.tex
\begin{tabular}{ c c c }
    \toprule
     \textbf{Relation} & \textbf{Path} & \textbf{Examples} \\ 
	 \midrule
     \multirow{2}{*}{\textsc{measure}} & [$w_2$] varies by [$w_1$] & \textit{state limit, age limit} \\  
      & 2,560 [$w_1$] portion of [$w_2$] & \textit{acre estate} \\ \hline
      \multirow{2}{*}{\textsc{personal title}} & [$w_2$] Anderson [$w_1$] & \textit{Mrs. Brown} \\ 
      & [$w_2$] Sheridan [$w_1$] & \textit{Gen. Johnson} \\ 
    \midrule
      \multirow{3}{*}{\textsc{create-provide-generate-sell}} & [$w_2$] produce [$w_1$] & \textit{food producer, drug group} \\  
      & [$w_2$] selling [$w_1$] & \textit{phone company, merchandise store} \\  
      & [$w_2$] manufacture [$w_1$] & \textit{engine plant, sugar company} \\ \hline  
      \multirow{2}{*}{\textsc{time-of1}} & [$w_2$] begin [$w_1$] & \textit{morning program} \\ 
      & [$w_2$] held Saturday [$w_1$] & \textit{afternoon meeting, morning session} \\ \hline
      \multirow{2}{*}{\textsc{substance-material-ingredient}}
      & [$w_2$] made of wood and [$w_1$] & \textit{marble table, vinyl siding} \\  
      & [$w_2$] material includes type of [$w_1$] & \textit{steel pipe} \\ 
	\bottomrule
\end{tabular}

%% file: analysis.tex
\begin{table*}[!t]
\small
\center
\input{figures/nc_embeddings_analysis.tex}
\vspace{-15pt}
\caption{Example of NCs from the \smalltt{Tratz-fine} random split test set, along with the most similar NC in the embeddings, where the two NCs have different labels.}
\label{tab:nc_embeddings}
\vspace{-12pt}
\end{table*}

\paragraph{Path Embeddings.} To focus on the changes from previous work, we analyze the performance of the path-based model on the \smalltt{Tratz-fine} random split. This dataset contains 37 relations and the model performance varies across them. Some relations, such as \textsc{measure} and \textsc{personal\_title} yield reasonable performance ($F_1$ score of 0.87 and 0.68). Table~\ref{tab:indicative_paths} focuses on these relations and illustrates the indicative paths that the model has learned for each relation. We compute these by performing the analysis in \newcite{shwartz-goldberg-dagan:2016:P16-1}, where each path is fed into the path-based model, and is assigned to its best-scoring relation. For each relation, we consider paths with a score $\geq 0.8$. 

Other relations achieve very low F$_1$ scores, indicating that the model is unable to learn them at all. Interestingly, the four relations with the lowest performance in our model\footnote{\textsc{lexicalized}, \textsc{topic\_of\_cognition\&emotion}, \textsc{whole+attribute\&feat}, \textsc{partial\_attr\_transfer}} are also those with the highest error rate in \newcite{W16-1604}, very likely since they express complex relations. For example, the \textsc{lexicalized} relation contains non-compositional NCs (\textit{soap opera}) or lexical items whose meanings departed from the combination of the constituent meanings. It is expected that there are no paths that indicate lexicalization. In \textsc{partial\_attribute\_transfer} (\textit{bullet train}), $w_1$ transfers an attribute to $w_2$ (e.g. \textit{bullet} transfers speed to \textit{train}). These relations are not expected to be expressed in text, unless the text aims to explain them (e.g. \textit{train as fast as a bullet}).

Looking closer at the model confusions shows that it often defaulted to general relations like \textsc{objective} (\textit{recovery plan}) or \textsc{relational-noun-complement} (\textit{eye shape}). The latter is described as ``indicating the complement of a relational noun (e.g., son of, price of)'', and the indicative paths for this relation indeed contain many variants of ``[$w_2$] of [$w_1$]'', which potentially can occur with NCs in other relations. The model also confused between relations with subtle differences, such as the different topic relations. Given that these relations were conflated to a single relation in the inter-annotator agreement computation in \newcite{tratz-hovy:2010:ACL}, we can conjecture that even humans find it difficult to distinguish between them.

\paragraph{NC Embeddings.} To understand why the NC embeddings did not contribute to the classification, we looked into the embeddings of the \smalltt{Tratz-fine} test NCs; 3091/3831 (81\%) of them had embeddings. 
For each NC, we looked for the 10 most similar NC vectors (in terms of cosine similarity), and compared their labels. We have found that only 27.61\% of the NCs were mostly similar to NCs with the same label. The problem seems to be inconsistency of annotations rather than low embeddings quality. Table~\ref{tab:nc_embeddings} displays some examples of NCs from the test set, along with their most similar NC in the embeddings, where the two NCs have different labels.

%% file: figures/nc_embeddings_analysis.tex
\begin{tabular}{ l l l l }
    \toprule
	\multicolumn{2}{c}{\textbf{Test NC}} & \multicolumn{2}{c}{\textbf{Most Similar NC}} \\ 
    \midrule
    \multicolumn{1}{c}{\textbf{NC}} & \multicolumn{1}{c}{\textbf{Label}} & \multicolumn{1}{c}{\textbf{NC}} & \multicolumn{1}{c}{\textbf{Label}}  \\ 
	\midrule
	\textit{majority party} & \smallsc{equative} & \textit{minority party} & \smallsc{whole+part\_or\_member\_of} \\ 
	\textit{enforcement director} & \smallsc{objective} & \textit{enforcement chief} & \smallsc{perform\&engage\_in} \\ 
	\textit{fire investigator} & \smallsc{objective} & \textit{fire marshal} & \smallsc{organize\&supervise\&authority} \\
	\textit{stabilization plan} & \smallsc{objective} & \textit{stabilization program} & \smallsc{perform\&engage\_in} \\ 
	\textit{investor sentiment} & \smallsc{experiencer-of-experience} & \textit{market sentiment} & \smallsc{topic\_of\_cognition\&emotion} \\ 
	\textit{alliance member} & \smallsc{whole+part\_or\_member\_of} & \textit{alliance leader} & \smallsc{objective} \\
	\bottomrule
\end{tabular}

%% file: conclusion.tex
We used an existing neural dependency path representation to represent noun-compound paraphrases, and along with distributional information applied it to the NC classification task. Following previous work, that suggested that distributional methods succeed due to lexical memorization, we show that when lexical memorization is not possible, the performance of all methods is much worse. Adding the path-based component helps mitigate this issue and increase performance. 

%% file: technical_details.tex
\section{Technical Details}
\label{sec:technical_details}

To extract paths, we use a concatenation of English Wikipedia and the Gigaword corpus.\footnote{\url{https://catalog.ldc.upenn.edu/ldc2003t05
}} We consider sentences with up to 32 words and  dependency paths with up to 8 edges, including satellites, and keep only 1,000 paths for each noun-compound. We compute the path embeddings in advance for all the paths connecting NCs in the dataset (\S\ref{sec:path_embeddings}), and then treat them as fixed embeddings during classification (\S\ref{sec:models}).

We use TensorFlow \cite{abadi2016tensorflow} to train the models, fixing the values of the hyper-parameters after performing preliminary experiments on the validation set. We set the mini-batch size to 10, use Adam optimizer \cite{kingma2014adam} with the default learning rate, and apply word dropout with probability 0.1. We train up to 30 epochs with early stopping, stopping the training when the $F_1$ score on the validation set drops 8 points below the best performing score.

We initialize the distributional embeddings with the 300-dimensional pre-trained GloVe embeddings \cite{pennington-socher-manning:2014:EMNLP2014} and the lemma embeddings (for the path-based component) with the 50-dimensional ones. Unlike HypeNET, we do not update the embeddings during training. The lemma, POS, and direction embeddings are initialized randomly and updated during training. NC embeddings are learned using a concatenation of Wikipedia and Gigaword. Similarly to the original GloVe implementation, we only keep the most frequent 400,000 vocabulary terms, which means that roughly 20\% of the noun-compounds do not have vectors and are initialized randomly in the model.